\documentclass[]{spie}  %>>> use for US letter paper
\usepackage[]{graphicx}
\usepackage{amssymb,amsmath}
\usepackage{url}

\title{Learning Hierarchical and Topographic Dictionaries \\ with Structured Sparsity}

\author{Julien Mairal\supit{a}, Rodolphe Jenatton\supit{b},
Guillaume Obozinski\supit{b} and Francis Bach\supit{b}
\skiplinehalf
\supit{a}Department of Statistics, University of California, Berkeley, USA. \\
\supit{b}INRIA - SIERRA Project-Team, Laboratoire d'Informatique de l'Ecole Normale Sup\'erieure (INRIA/ENS/CNRS UMR 8548), 23, avenue d'Italie 75214 Paris CEDEX 13, France.
}

\authorinfo{J.M.: E-mail: julien@stat.berkeley.edu\\  R.J., G.O., F.B.: E-mail: firstname.lastname@inria.fr}
\def\x{{\mathbf x}}
\def\z{{\mathbf z}}
\def\1{{\mathbf 1}}

\def\X{{\mathbf X}}

\def\betab{{\boldsymbol\beta}}

\def\alphab{{\boldsymbol\alpha}}

\def\y{{\mathbf y}}
\def\w{{\mathbf w}}
\def\D{{\mathbf D}}
\def\Y{{\mathbf Y}}

\def\d{{\mathbf d}}
\def\E{{\mathbb E}}

\def\C{{\mathcal C}}
\def\GG{{\mathcal G}}

\def\L{{\mathcal L}}

\def\d{{\mathbf d}}

\def\u{{\mathbf u}}

\def\Real{{\mathbb R}}

\def\u{{\mathbf u}}
\def\A{{\mathbf A}}

\def\sign{\operatorname{sign}}
\def\st{~~\text{s.t.}~~}
\def\defin{\triangleq}

\newcommand{\BlackBox}{\rule{1.5ex}{1.5ex}}
\newcommand{\R}[1]{\mathbb{R}^{#1}}

\newcommand{\G}{\mathcal{G}}

\newcommand{\NormDeux}[1]{\left\|#1\right\|_2}

\newcommand{\NormFro}[1]{\left\|#1\right\|_{\text{F}}}

\newcommand{\IntSet}[1]{[ 1;#1 ]}%{\textlbrackdbl #1 \textrbrackdbl}%{\text{\textlbrackdbl} #1 \text{\textrbrackdbl}}
\newcommand{\InSet}[1]{[ 1;#1 ]}%{\textlbrackdbl #1 \textrbrackdbl}%{\text{\textlbrackdbl} #1 \text{\textrbrackdbl}}

\long\def\symbolfootnote[#1]#2{\begingroup\def\thefootnote{\fnsymbol{footnote}}\footnote[#1]{#2}\endgroup} 
\renewcommand{\w}{\alphab}
\newcommand{\citet}{\cite}
\newcommand{\citep}{\cite}
\renewcommand{\u}{\betab}
\renewcommand{\z}{\betab}
\renewcommand{\Omega}{\psi}
  
\begin{document} 
  \maketitle 

%%%%%%%%%%%%%%%%%%%%%%%%%%%%%%%%%%%%%%%%%%%%%%%%%%%%%%%%%%%%% 
\begin{abstract}
Recent work in signal processing and statistics have focused on
defining new regularization functions, which not only induce sparsity of
the solution, but also take into account the structure of the
problem~\cite{zhao,jenatton,jacob,huang,baraniuk,cehver,he}. We present in this
paper a class of convex penalties introduced in the machine learning
community, which take the form of a sum of~$\ell_2$-
and~$\ell_\infty$-norms over groups of variables. They extend the classical
group-sparsity regularization\cite{turlach,yuan,obozinski} in the sense that the groups
possibly overlap, allowing more flexibility in the group design.
We review efficient optimization methods to deal with the corresponding inverse
problems~\cite{jenatton4,mairal10,mairal11}, and their application to the
problem of learning dictionaries of natural image
patches~\cite{field,field2,engan,elad,mairal7}: On the one hand, dictionary
learning has indeed proven effective for various signal processing
tasks~\cite{elad,mairal}.  On the other hand, structured
sparsity provides a natural framework for modeling dependencies between
dictionary elements.  We thus consider a structured sparse regularization
to learn dictionaries embedded in a particular structure, for instance a
tree~\cite{jenatton4} or a two-dimensional grid~\cite{kavukcuoglu2}. In the latter
case, the results we obtain are similar to the dictionaries produced by
topographic independent component analysis~\cite{hyvarinen2}.
\end{abstract}


%%%%%%%%%%%%%%%%%%%%%%%%%%%%%%%%%%%%%%%%%%%%%%%%%%%%%%%%%%%%%
\section{INTRODUCTION}

Sparse representations have recently drawn much interest in signal,
image, and video processing. Under the assumption that natural
images admit a sparse decomposition in some redundant basis (or so-called
{\em dictionary}), several such models have been proposed, e.g.,
curvelets\cite{candes}, wedgelets\cite{donoho}, bandlets~\cite{mallat3}
and more generally various sorts of wavelets\cite{mallat}. 
Learned sparse image models were first introduced in the
neuroscience community by Olshausen and Field\cite{field,field2} for
modeling the spatial receptive fields of simple cells in the mammalian
visual cortex. The linear decomposition of a
signal using a few atoms of a \emph{learned} dictionary instead of
predefined ones, has recently led to state-of-the-art
results for numerous low-level image processing tasks such as denoising, inpainting
\cite{elad,mairal,mairal8} or texture synthesis \cite{peyre}, showing
that sparse models are well adapted to natural images. Unlike
decompositions based on principal component analysis, these models can
rely on overcomplete dictionaries, with a number of atoms greater than
the original dimension of the signals, allowing more flexibility to adapt
the representation to the data.

In addition to this recent interest from the signal and image processing
communities for sparse modelling, statisticians have
developed similar tools from a different point of view. In signal
processing, one often represents a data vector~$\y$ of fixed 
dimension~$m$ as a linear combination of~$p$ dictionary
elements~$\D=[\d^1,\ldots,\d^p]$ in~$\Real^{m \times p}$. In other words,
one looks for a vector~$\alphab$ in~$\Real^p$ such that~$\y \approx
\D\alphab$.  When we assume~$\alphab$ to be sparse---that is has a lot of
zero coefficients, we obtain a sparse linear model and need appropriate
regularization functions.  When~$\D$ is fixed, the columns~$\d^i$
can be interpreted as the elements of a redundant basis, for instance
wavelets~\cite{mallat}.

Let us now consider a different problem occurring in statistics or machine
learning. Given a training set~$(y^i,\x^i)_{i=1}^m$, where the~$y^i$'s are
scalars, and the~$\x^i$'s are vectors in~$\Real^p$, the task is to predict a
value for~$y$ from an observation~$\x$ in~$\Real^p$.  This is usually
achieved by \emph{learning} a model from the training data, and the simplest
one is to assume that there exists a linear relationship $y \approx
\x^\top{\mathbf w}$, where~${\mathbf w}$ is a vector in~$\Real^p$. Learning the model
amounts to adapt~${\mathbf w}$ to the training set and denoting by~$\y$ the vector
in~$\Real^n$ whose entries are the $y^i$'s, and~$\X$ the matrix in~$\Real^{m
\times p}$ the matrix whose rows are the~$\x^i$'s, we end up looking for a
vector~${\mathbf w}$ such that~$\y \approx \X{\mathbf w}$.  When one knows in advance
that the vector~${\mathbf w}$ is sparse, a similar problem as in signal
processing is raised, where~$\X$ can be interpreted as a ``dictionary'' but is often
called a set of ``features'' or ``predictors''.
This is therefore not surprising that both communities have developed similar
tools, the Lasso formulation~\cite{tibshirani}, L2-boosting
algorithm\cite{friedman2001}, forward selection techniques\cite{weisberg} in
statistics are respectively equivalent (up to minor details) to the basis
pursuit problem~\cite{chen}, matching and variants of orthogonal matching pursuit
algorithms\cite{mallat4}.

Formally, the sparse decomposition problem of a signal~$\y$ using a
dictionary~$\D$ amounts to finding a vector~$\alphab$ minimizing the following
cost function
\begin{equation}
   \min_{\alphab \in \Real^{p}} \frac{1}{2}\|\y-\D\alphab\|_2^2 + \lambda\psi(\alphab), \label{eq:lasso}
\end{equation}
where~$\psi$ is a sparsity-inducing function, and~$\lambda$ a regularization parameter.
A natural choice is to use the~$\ell_0$ quasi-norm, which counts the number of non-zero
elements in a vector, leading however to an NP-hard problem\cite{natarajan}, which is usually tackled with greedy algorithms~\cite{mallat4}.
Another approach consists of using a convex relaxation such 
as the~$\ell_1$-norm. Indeed, it is well known that the $\ell_1$ penalty
yields a sparse solution, but there is no analytic link between the value
of $\lambda$ and the effective sparsity $\|\x\|_0$ that it yields.

We consider in this paper recent sparsity-inducing penalties
capable of encoding the structure of a signal decomposition on a redundant basis.  The
$\ell_1$-norm primarily encourages sparse solutions, regardless of the
potential structural relationships (e.g., spatial, temporal or hierarchical)
existing between the variables.  To cope with that issue,
some effort has recently been devoted to
designing sparsity-inducing regularizations capable of encoding higher-order
information about the patterns of non-zero
coefficients, some of these works coming from the machine learning/statistics 
literature~\cite{jenatton,jacob,zhao,huang} others from signal processing \cite{baraniuk}.
We use here the approach of Jenatton et al.~\cite{jenatton} who 
consider sums of norms of appropriate
subsets, or \textit{groups}, of variables,  in order to control the
sparsity patterns of the solutions.  The underlying optimization is usually
difficult, in part because it involves nonsmooth components. We review strategies
to address these problems, first when the groups are embedded in a
tree~\cite{zhao,jenatton4}, second in a general setting~\cite{mairal11}.

Whereas these penalties have been shown to be useful for solving various
problems in computer vision, bio-informatics, or
neuroscience~\cite{jenatton,jacob,zhao,huang}, we address here the problem of
learning dictionaries of natural image patches which exhibit particular
relationships among their elements.  Such a construction is motivated a priori
by two distinct but related goals: first to potentially improve the performance
of denoising, inpainting or other signal processing tasks that can be tackled
based on the learned dictionaries, and second to uncover or reveal some of the
natural structures present in images. 
In previous work\cite{jenatton4}, we have for instance embedded
dictionary elements into a tree, by using a hierarchical norm~\cite{zhao}.
This model encodes a rule saying that a dictionary element can be used in the
decomposition of a signal only if its ancestors in the tree are used as well,
similarly as in the zerotree wavelet model~\cite{shapiro2}.
In the related context of independent component analysis (ICA), Hyv\"arinen et
al.\cite{hyvarinen2} have arranged independent components (corresponding to
dictionary elements) on a two-dimensional grid, and have modelled spatial
dependencies between them. When learned on whitened natural image patches, this
model exhibits ``Gabor-like'' functions which are smoothly organized on the
grid, which the authors call a topographic map.  As shown
in Ref.~\citenum{kavukcuoglu2}, such a result can be reproduced with a dictionary
learning formulation using structured regularization.

We use the following notation in the paper: Vectors are denoted by bold lower
case letters and matrices by upper case ones.  We define for $q \geq 1$ the
\mbox{$\ell_q$-norm} of a vector~$\x$ in~$\Real^m$ as $\|\x\|_q \defin
(\sum_{i=1}^m |\x_i|^q)^{{1}/{q}}$, where~$\x_i$ denotes the $i$-th coordinate
of~$\x$, and $\|\x\|_\infty \defin \max_{i=1,\ldots,m} |\x_i| = \lim_{q \to
\infty} \|\x\|_q$.  We also define the $\ell_0$-pseudo-norm as the number of
nonzero elements in a vector:\footnote{Note that it would be more proper to
write $\|\x\|_0^0$ instead of $\|\x\|_0$ to be consistent with the traditional
notation $\|\x\|_q$.  However, for the sake of simplicity, we will keep this
notation unchanged in the rest of the paper.} $\|\x\|_0 \defin \#\{i \st \x_i
\neq 0  \} = \lim_{q \to 0^+}  (\sum_{i=1}^m |\x_i|^q)$.  We consider the
Frobenius norm of a matrix~$\X$ in~$\Real^{m \times n}$: $\NormFro{\X} \defin
(\sum_{i=1}^m \sum_{j=1}^n \X_{ij}^2)^{{1}/{2}}$, where $\X_{ij}$ denotes the
entry of~$\X$ at row $i$ and column $j$. 

This paper is structured as follows: Section~\ref{sec:related} reviews the
dictionary learning and structured sparsity frameworks,
Section~\ref{sec:optim} is devoted to optimization techniques, and
Section~\ref{sec:exp} to experiments with structured dictionary learning.
Note that the material of this paper relies upon two of our papers published in
the Journal of Machine Learning Research~\cite{mairal11,jenatton4}. 

\section{RELATED WORK}\label{sec:related}
We present in this section the dictionary learning framework and
structured sparsity-inducing regularization functions.
\subsection{Dictionary Learning}\label{subsec:dict}
Consider a signal $\y$ in $\Real^m$. We say that $\y$ admits
a sparse approximation over a dictionary $\D$ in $\Real^{m \times
p}$, composed of $p$ elements (atoms), when we can find a linear
combination of a ``few'' atoms from $\D$ that is ``close'' to the
original signal $\x$. A number of practical algorithms have been developed for
learning such dictionaries like the K-SVD algorithm \cite{aharon}, the method
of optimal directions (MOD) \cite{engan}, stochastic gradient descent
algorithms~\cite{field} or other online learning
techniques~\cite{mairal7}, which will be briefly reviewed in Section~\ref{sec:optim}.
This approach has led to several restoration algorithms, with state of the art
results in image and video denoising, inpainting, demosaicing
\cite{elad,mairal}, and texture synthesis \cite{peyre}.

Given a training set of signals $\Y=[\y^1,\ldots,\y^n]$ in~$\Real^{m \times
n}$, such as natural image patches, dictionary learning amounts to finding a
dictionary which is adapted to every signal~$\y^i$, in other words it can be
cast as the following optimization problem
\begin{equation}
   \min_{\D \in \C, \A \in \Real^{p \times n}} \sum_{i=1}^n
\frac{1}{2}\|\y^i-\D\alphab^i\|_2^2 + \lambda\psi(\alphab^i),\label{eq:dict}
\end{equation}
where~$\A=[\alphab^1,\ldots,\alphab^n]$ are decomposition coefficients, $\psi$
is a sparsity-inducing penalty, and~$\C$ is a constraint set, typically the set
of matrices whose columns have less than unit~$\ell_2$-norm:
\begin{equation}
\C \defin \{ \D \in \Real^{m \times p} :  \forall j=1,\ldots,p,~~~~\|\d^j\|_2 \leq 1 \}.
\end{equation}
To prevent $\D$ from being arbitrarily large (which would lead to arbitrarily
small values of $\alphab$), it is indeed necessary to constrain the dictionary
with such a set~$\C$. 
We also remark that dictionary learning is an instance of matrix factorization
problem, which can be equivalently rewritten
\begin{equation}
  \min_{\D \in \C, \A \in \Real^{p \times n}} \frac{1}{2}\|\Y-\D\A\|_{\text{F}}^2 + \lambda\psi'(\A), \label{eq:dict2}
\end{equation}
with an appropriate function~$\psi'$. Noticing this interpretation 
of dictionary learning as a matrix factorization has a number of practical
consequences. With adequate constraints on~$\A$ and~$\D$, one can indeed recast
several classical problems as regularized matrix factorization
problems, for instance principal component analysis (PCA), non-negative matrix
factorization (NMF)\cite{lee2}, hard and soft vector quantization (VQ). As a
first consequence, \emph{all of these approaches can be addressed with similar
algorithms, as shown in Ref.~\citenum{mairal7}}.  A natural approach to approximately solve this non-convex problem is for instance to alternate
between the optimization of~$\D$ and~$\A$ in Eq.~(\ref{eq:dict2}),
minimizing over one while keeping the other one fixed\cite{engan}, a technique
also used in the K-means algorithm for vector quantization.

Another approach consists of using stochastic approximations and use online
learning algorithms. When~$n$ is large, finding the sparse coefficients~$\A$
with a fixed dictionary~$\D$ requires solving~$n$ sparse decomposition
problems~(\ref{eq:lasso}), which can be cumbersome. To cope with this issue,
online learning techniques adopt a different iterative algorithmic scheme: At
iteration~$t$, they randomly draw one signal~$\y^t$ from the training set (or a
mini-batch), and try to ``improve'' $\D$ given this observation.
Assume indeed that~$n$ is large and that the image patches~$\y^i$ are i.i.d.
samples drawn from an unknown distribution~$p(\y)$, then Eq.~(\ref{eq:dict}) is
asymptotically equivalent to
\begin{equation}
   \min_{\D \in \C} \E_{\y\sim p(\y)} \left[\min_{\alphab \in \Real^p}
\frac{1}{2}\|\y-\D\alphab\|_2^2 + \lambda\psi(\alphab)\right].\label{eq:dict3}
\end{equation}
In order to optimize a cost function which includes an expectation, it
is natural to use stochastic approximations\cite{kushner}.
When~$\psi$ is the~$\ell_1$-norm, this problem is also
under mild assumptions differentiable (see Mairal et al.\cite{mairal7}
for more details), and a first order stochastic gradient descent
step\cite{field2,mairal7}, given a signal~$\y^t$ can
can be written: 
\begin{equation}
   \D \leftarrow \Pi_\C\Big[\D + \delta_t(\y^t-\D\alphab^t)\alphab^{t\top})\Big], \label{eq:sgd}
\end{equation}
where $\delta_t$ is the gradient step, $\Pi_\C$ is the orthogonal projector
onto $\C$. The vector~$\alphab^t$ carries the sparse coefficients obtained from
the decomposition of~$\y^t$ with the current dictionary~$\D$. When~$\psi$ is
the~$\ell_0$-norm, this iteration is heuristic but gives good results in practice,
when~$\psi$ is the~$\ell_1$-norm, and assuming the solution of the sparse decomposition
problem to be unique, this iteration exactly corresponds to a stochastic gradient descent
algorithm\cite{kushner}.  Note that the vectors~$\y^t$ are assumed to be i.i.d. samples of
the (unknown) distribution $p(\y)$. Even though it is often difficult to obtain
such i.i.d. samples, the vectors~$\y^t$ are in practice obtained by cycling on
a randomly permuted training set.  The main difficulty in this approach
is to take a good learning rate~$\delta_t$.
Other dedicated online learning algorithms have been proposed~\cite{mairal7},
which can be shown to provide a stationary point of the optimization problem~(\ref{eq:dict3}).
All of these online learning techniques have shown to yield significantly speed-ups over classical
alternative minimization approach, when $n$ is large enough.

\begin{figure}[hbtp]
\centering
\includegraphics[width=0.45\linewidth]{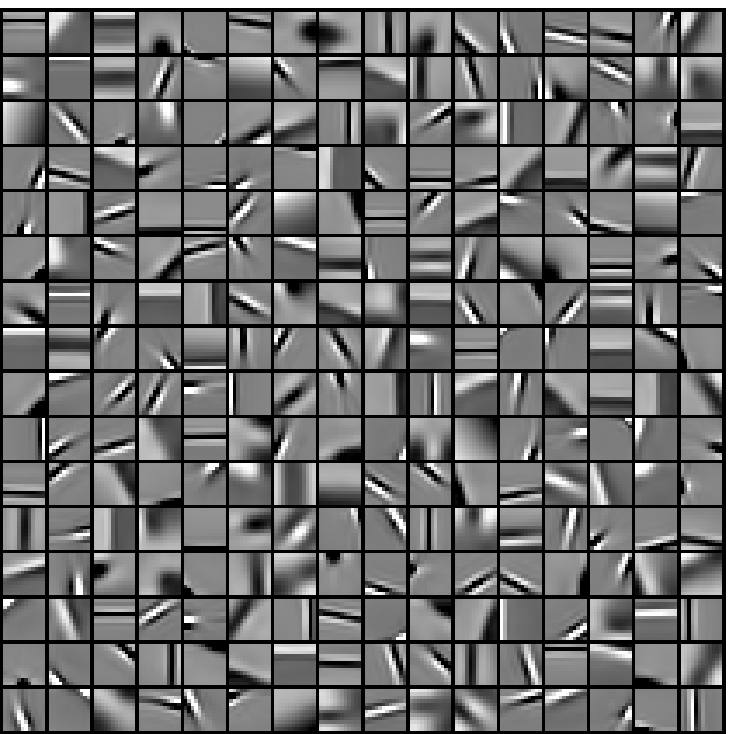}
\hfill
\includegraphics[width=0.45\linewidth]{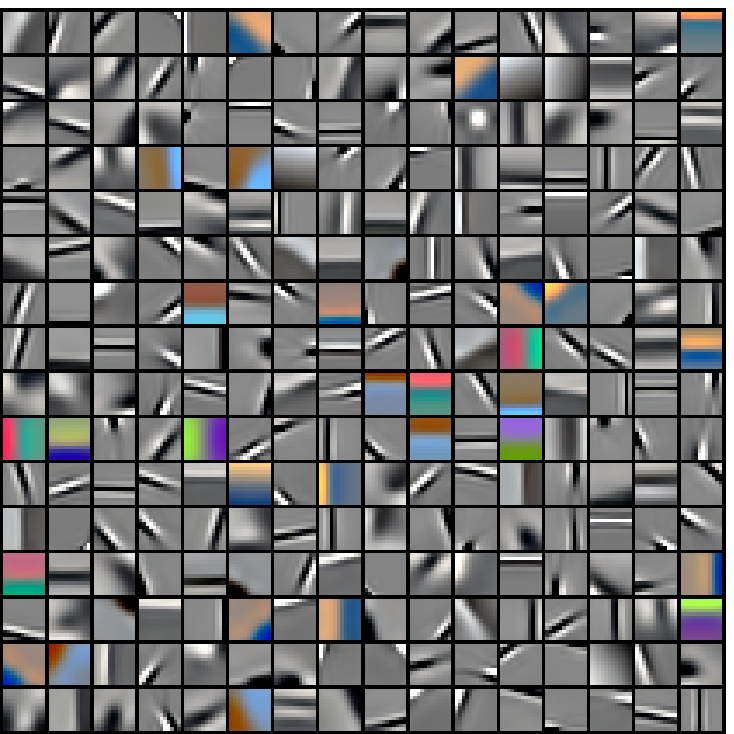}
\caption{Examples of dictionary with $p=256$ elements, learned on a
database of $10$ million natural $12 \times 12$ image patches when~$\psi$
is the~$\ell_1$-norm, for grayscale patches on the left, and color
patches in the right (after removing the mean color of each
patch).  Image taken from Ref.~\citenum{mairal_thesis}.}\label{fig:patches}
\end{figure}

Examples of dictionaries learned using the approach of Mairal et al.\cite{mairal7} are
represented in Figure~\ref{fig:patches}, and exhibit intriguing visual
results. Some of the dictionary elements look like Gabor wavelets,
whereas other elements are more difficult to interpret. As for the color
image patches, we observe that most of the dictionary elements are gray,
with a few low-frequency colored elements exhibiting complementary colors, a phenomenon already observed in image processing applications~\cite{mairal}.

\subsection{Structured Sparsity}
We consider again the sparse decomposition problem presented in
Eq.~(\ref{eq:lasso}), but we allow~$\psi$ to be different than
the~$\ell_0$ or~$\ell_1$-regularization, and we are interested in
problems where the solution is beforehand not only assumed to be sparse
---that is, the solution has only a few non-zero coefficients,
but also to form non-zero patterns with a specific structure.
It is indeed possible to encode
additional knowledge in the regularization other than just sparsity. 
For instance, one may want the
non-zero patterns to be structured in the form of non-overlapping
groups~\cite{turlach,yuan,obozinski}, in a tree~\cite{zhao,jenatton4},
or in overlapping groups~\cite{jenatton,jacob,huang,baraniuk,cehver,he}.
As for classical non-structured sparse models, there are basically two
lines of research, that either (a) deal with nonconvex and combinatorial
formulations that are in general computationally intractable and
addressed with greedy algorithms or (b) concentrate on convex relaxations
solved with convex programming methods. We focus in this paper on the latter.

When the sparse coefficients are organized in groups, a penalty encoding
explicitly this prior knowledge can improve the prediction performance
and/or interpretability of the learned models~\cite{yuan,obozinski}. Denoting by~$\GG$ a set of groups of indices, such a penalty takes the~form:
\begin{equation}
   \psi(\alphab) \, \defin\, \sum_{g \in \GG} \eta_g \|\alphab_g\|_q, \label{eq:def_omega}
\end{equation}
where $\alphab_j$ is the $j$-th entry of $\alphab$ for $j$ in
$\IntSet{p}\defin\{1,\ldots,p\}$, the vector $\alphab_g$ in $\R{|g|}$ records
the coefficients of $\w$ indexed by $g$ in $\GG$, and the scalars
$\eta_g$ are positive weights. $\|.\|_q$ denotes here either the~$\ell_2$ or~$\ell_\infty$-norms.
Note that when $\GG$ is the set of singletons of $\IntSet{p}$, we get
back the $\ell_1$-norm.
Inside a group, the~$\ell_2$- or~$\ell_\infty$-norm does not induce sparsity, whereas the sum over the groups can be interpreted
as an~$\ell_1$-norm\footnote{The sum of positive values is equal to the~$\ell_1$-norm of a vector carrying these values.}
and indeed, when $\GG$ is a \emph{partition} of $\IntSet{p}$, variables are selected in
groups rather than individually.  When the groups overlap, $\Omega$ is
still a norm and sets groups of variables to zero
together~\cite{jenatton}.  The latter setting has first been considered
for hierarchies~\cite{zhao}, and then extended to general group
structures~\cite{jenatton}.\footnote{Note that
other sparsity inducing norms have been introduced\cite{jacob}, which are different and not equivalent to the one we consider in this paper. One should be careful when referring to ``structured sparsity penalty with overlapping groups'', since different generalizations of the selection of variable in groups have been proposed.}  Solving Eq.~(\ref{eq:lasso}) in this context becomes challenging and is the topic of the next section. Before that, in order to better illustrate how such norms should be used and how to design a group structure inducing a desired sparsifying effect, we proceed by giving a few examples of group structures.

\subsubsection{One-dimensional Sequence.}
\begin{figure}[!h]
\begin{center}
\includegraphics[scale=.55]{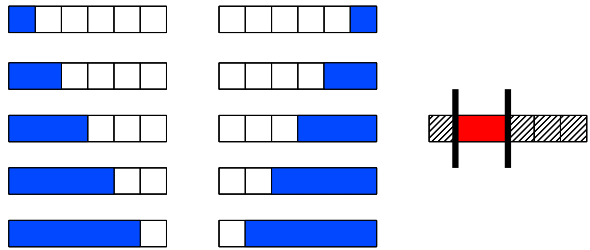}
\end{center}
\caption{ (Left) The set of blue groups to penalize in order to select contiguous patterns in a sequence.
(Right) In red, an example of such a nonzero pattern with its corresponding zero pattern (hatched area). Image taken from Ref.~\citenum{jenatton}.} 
\label{intro:fig:sequence}
 \end{figure}

Given $p$ variables organized in a sequence, suppose we want to select
only contiguous nonzero patterns. A set of groups~$\G$ exactly producing
such patterns is represent on Figure~\ref{intro:fig:sequence}. It is indeed
easy to show that by selecting a family of groups in~$\G$ represented in this figure, and
setting the corresponding variables to zero, exactly leads to contiguous
patterns of non-zero coefficients. The penalty~(\ref{eq:def_omega}) with
this group structure produces therefore exactly the desired sparsity patterns.
\subsubsection{Hierarchical Norms}
\begin{figure}[hbtp]
   \centering
   \includegraphics[width=0.6\textwidth]{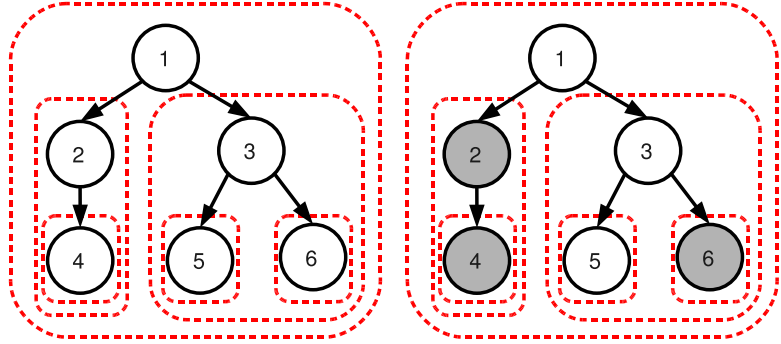}
   \caption{Left: example of a tree-structured set of groups $\G$ (dashed contours in red), corresponding to a tree $\mathcal{T}$ with $p=6$ nodes represented by black circles.
   Right: example of a sparsity pattern induced by the tree-structured norm corresponding to $\G$: the groups $\{2,4\},\{4\}$ and $\{6\}$ are set to zero, so that the corresponding nodes (in gray) that form subtrees of $\mathcal{T}$ are removed.
   The remaining nonzero variables $\{1,3,5\}$ form a rooted and connected subtree of $\mathcal{T}$.
   This sparsity pattern obeys the following equivalent rules: (i) if a node is selected, so are all of its ancestors. (ii) if a node is not selected, then its descendant are not selected. Image taken from Ref.~\citenum{jenatton4}.
}\label{fig:tree}
\end{figure}

Another example of interest originally comes from the wavelet literature.
It consists of modelling hierarchical relations between wavelet coefficients,
which are naturally organized in a tree, due to the multiscale properties of
wavelet decompositions.\cite{mallat} The zero-tree wavelet
model\cite{shapiro2} indeed assumes that if a wavelet coefficient is set
to zero, then it should be the case for all its descendants in the tree.
This effect can in fact be exactly achieved with the convex regularization
of Eq.~(\ref{eq:def_omega}), with an appropriate group structure
presented in Figure~\ref{fig:tree}. This penalty was originally introduced in the statistics community by Zhao et al.\cite{zhao}, and found different applications, notably in topic models for text corpora~\cite{jenatton4}.
\subsubsection{Neighborhoods on a 2D-Grid}
Another group structure we are going to consider corresponds to the assumption that the dictionary elements
can be organized on a 2D-grid, for example we might
have~$p=20\times 20$ dictionary elements. To obtain a spatial
regularization effect on the grid, it is possible to use as groups all
the~$e \times e$ neighborhoods on the grid, for example~$3 \times 3$.
The main effect of such a regularization is to encourage variables that
are in a same neighborhood to be set to zero all together. Such 
dictionary structure has been used for instance in Ref.~\citenum{mairal11}
for a background subtraction task (segmenting foreground objects from the
background in a video).

\section{OPTIMIZATION FOR STRUCTURED SPARSITY}\label{sec:optim}
We now present optimization techniques to solve Eq.~(\ref{eq:lasso}) when~$\psi$ is a structured norm~(\ref{eq:def_omega}). This is the main difficulty to overcome to learn structured dictionaries.
We review here the techniques introduced in Refs.~\citenum{jenatton4,mairal11}. More details can be found in these two papers. Other technique for
dealing with sparsity-inducing penalties can also be found in Ref.~\citenum{bach5}.
\subsection{Proximal Gradient Methods}\label{subsec:optim1}
In a nutshell, proximal methods can be seen as a natural extension of
gradient-based techniques, and they are well suited to minimizing the sum $f+\lambda \psi$ of two convex
terms, a smooth function~$f$ ---continuously differentiable with Lipschitz-continuous gradient--- and a potentially non-smooth function~$\lambda \psi$ (see Refs.~\citenum{combette,bach5} and references therein). In our context, the function~$f$ takes the form~$f(\alphab)=\frac{1}{2}\|\y-\D\alphab\|_2^2$.
At each iteration, the function $f$ is linearized at the
current estimate~$\w_0$ and the so-called
\textit{proximal} problem has to be solved:
\begin{displaymath}
    \min_{\w \in \R{p}} f(\w_0) + (\w - \w_0)^\top \nabla f(\w_0) + \lambda \Omega(\w) + {\displaystyle \frac{L}{2}}\|\w - \w_0\|_2^2.
\end{displaymath}
The quadratic term keeps the solution in a neighborhood where the current linear
approximation holds, and $L\!>\!0$ is an upper bound on the Lipschitz
constant of $\nabla f$. This problem can be rewritten~as
\begin{equation}\label{eq:prox_problem}
   \min_{\w \in \R{p}} {\displaystyle \frac{1}{2}} \NormDeux{\u-\w}^2 + \lambda'  \Omega(\w),
\end{equation}
with $\lambda' \defin \lambda / L$, and $\u \defin \w_0 - \frac{1}{L} \nabla f(\w_0)$.
We call \textit{proximal operator} associated with the regularization
$\lambda'\Omega$ the function that maps a vector~$\u$ in~$\R{p}$ onto the
(unique, by strong convexity) solution~$\w^{\star}$ of Eq.~(\ref{eq:prox_problem}). 
Simple proximal methods use $\w^{\star}$ as the next iterate, but accelerated variants~\cite{nesterov,beck} are also based on the proximal operator
and require to solve problem (\ref{eq:prox_problem}) efficiently to enjoy their fast convergence rates.

This has been shown to be possible in many cases:
\begin{itemize}
   \item When $\Omega$ is the $\ell_1$-norm---that is $\Omega(\w)=\|\w\|_1$---
      the proximal operator is the well-known elementwise soft-thresholding
      operator,
      \begin{displaymath}
         \forall j \in \InSet{p},~~\u_j \mapsto \sign(\u_j)(|\u_j|-\lambda)_+~~~ = \begin{cases}
            0 & \text{if}~~ |\u_j| \leq \lambda \\
            \sign(\u_j)(|\u_j|-\lambda) & \text{otherwise}. 
         \end{cases}
      \end{displaymath}
   \item When $\Omega$ is a group-Lasso penalty with $\ell_2$-norms---that is,
      $\Omega(\u)=\sum_{g\in \GG}\|\u_g\|_2$, with $\GG$ being a partition of
      $\InSet{p}$, the proximal problem is \emph{separable} in every group, 
      and the solution is a generalization of the soft-thresholding operator to
      groups of variables:
      \begin{displaymath}
         \forall g\in \G~~, \u_g \mapsto \u_g-\Pi_{\|.\|_2 \leq \lambda}[\u_g] = \begin{cases}
            0 & \text{if}~~ \|\u_g\|_2 \leq \lambda \\
	    \frac{\|\u_g\|_2-\lambda}{\|\u_g\|_2}\u_g & \text{otherwise}, \\
         \end{cases} 
      \end{displaymath}
      where $\Pi_{\|.\|_2 \leq \lambda}$ denotes the orthogonal projection onto the ball of 
      the $\ell_2$-norm of radius $\lambda$.
   \item When $\Omega$ is a group-Lasso penalty with $\ell_\infty$-norms---that is,
      $\Omega(\u)=\sum_{g\in \G}\|\u_g\|_\infty$, with $\GG$ being a partition of
      $\InSet{p}$, the solution is a different group-thresholding
      operator:
      \begin{displaymath}
         \forall g\in \G,~~\u_g \mapsto \u_g - \Pi_{\|.\|_1 \leq \lambda}[\u_g],
      \end{displaymath}
      where $\Pi_{\|.\|_1 \leq \lambda}$ denotes the orthogonal projection onto
      the $\ell_1$-ball of radius $\lambda$, which can be solved in $O(p)$
      operations~\cite{brucker,maculan}. Note that when $\|\u_g\|_1 \leq \lambda$, we
      have a group-thresholding effect, with $\u_g - \Pi_{\|.\|_1 \leq \lambda}[\u_g]=0$.
  \item When $\Omega$ is a tree-structured sum of $\ell_2$- or $\ell_\infty$-norms as introduced by~Ref.~\citenum{zhao}---meaning that two groups are either disjoint or one is included in the other, the solution admits a closed form. 
Let $\preceq$ be a total order on $\G$ such that for $g_1, g_2$ in $\G$, $g_1 \preceq g_2$ if and only if either $g_1 \subset g_2$ or $g_1 \cap g_2=\emptyset$.\footnote{For a tree-structured set $\G$, such an order exists.} Then, if $g_1 \preceq \ldots \preceq g_{|\G|}$, and if we define $\text{Prox}^g$ as (a) the proximal operator $\u_g \mapsto \text{Prox}_{\lambda \eta_g \|\cdot\|}(\u_g)$ on the subspace corresponding to group $g$ and (b) the identity on the orthogonal, it is shown in Ref.~\citenum{jenatton4} that:
\begin{equation}
\text{Prox}_{\lambda\Omega}=\text{Prox}^{g_m} \circ \ldots \circ \text{Prox}^{g_1},
\end{equation}
which can be computed in $O(p)$ operations. It also includes the sparse group Lasso (sum of group-Lasso penalty and $\ell_1$-norm) of~Refs.~\citenum{friedman2} and \citenum{sprechmann}.
 \item When the groups overlap but do not have a tree structure, computing the proximal operator is more difficult, but it can still be done efficiently when $q=\infty$.
Indeed, as shown by Mairal et al.~\cite{mairal10}, there exists a dual relation between such an operator and a quadratic min-cost flow problem on a particular graph, which can be tackled using network flow optimization techniques. Moreover, it may be extended to more general situations where structured sparsity is expressed through submodular functions~\cite{bach2010structured}.
\end{itemize}
Mainly using the tools of Refs.~\citenum{jenatton4,mairal10}, we are therefore able
to efficiently solve Eq.~(\ref{eq:lasso}), either in the case of hierarchical
norms with~$\ell_2$- or~$\ell_\infty$-norms, or with general group structures
with~$\ell_\infty$-norms.  This is one of the main requirements to be able to
learn structured dictionary.  The next section presents a different
optimization technique, adapted to any group structure 
with~$\ell_2$- or~$\ell_\infty$-norms.

\subsection{Augmenting Lagrangian Techniques}\label{subsec:optim2}
We consider a class of algorithms which leverage the concept of variable
splitting~\cite{combette,bertsekas3,tomioka,boyd2}.
The key is to introduce additional variables $\z^g$ in $\Real^{|g|}$, one for every group $g$ in $\GG$,
and equivalently reformulate Eq.~(\ref{eq:lasso}) as
\begin{equation}
   \min_{\substack{ \w \in \Real^p\\ \z^g \in \Real^{|g|}\ \text{for}\ g \in \GG} } f(\w) + \lambda
\sum_{g \in \GG} \eta_g \|\z^g\|_q   \st  \forall g \in \GG,~~ \z^g = \w_g, \label{eq:varsplit}
\end{equation}
The issue of overlapping groups is removed, but new constraints and variables are added.

To solve this problem, it is possible to use the so-called alternating direction method of
multipliers (ADMM)~\cite{combette,bertsekas3,tomioka,boyd2}.\footnote{This
method is used in Ref.~\citenum{sprechmann} for computing the
proximal operator associated to hierarchical norms, and in the same context as ours in Refs.~\citenum{boyd2} and~\citenum{qin}.}
It introduces dual variables $\nu^g$ in $\Real^{|g|}$ for all $g$ in $\GG$, and defines the
augmented Lagrangian:
\begin{displaymath}
   \L\big(\w,(\z^g)_{g\in\G},(\nu^g)_{g\in\G}\big) \defin f(\w) + \sum_{g \in \GG} 
   \big[\lambda\eta_g \|\z^g\|  + \nu^{g\top}(\z^g-\w_g) + \frac{\gamma}{2} \|\z^g - \w_g \|_2^2\big],
\end{displaymath}
where $\gamma > 0$ is a parameter.
It is easy to show that solving Eq.~(\ref{eq:varsplit}) amounts to finding a
saddle-point of the augmented Lagrangian.\footnote{The augmented Lagrangian is
in fact the classical Lagrangian~\cite{boyd} of the following
optimization problem which is equivalent to Eq.~(\ref{eq:varsplit}):
\begin{displaymath}
   \min_{\w \in \Real^p, (\z^g \in \Real^{|g|})_{g \in \GG} } f(\w) + \lambda
\sum_{g \in \GG} \eta_g \|\z^g\| + \frac{\gamma}{2}\|\z^g-\w_g\|_2^2  \st  \forall g \in \GG,~~ \z^g = \w_g.
\end{displaymath}
}
The ADMM algorithm finds such a saddle-point by iterating between the minimization of $\L$ with respect to each primal variable, keeping the other ones fixed,
and gradient ascent steps with respect to the dual variables. More precisely, it can be summarized~as: %the following iterative procedure is used:
\begin{enumerate}
   \item Minimize $\L$ with respect to $\w$, keeping the other variables fixed. \label{admm:step1}
   \item Minimize $\L$ with respect to the $\z^g$'s, keeping the other variables fixed. The solution can be obtained in closed form: for all $g$ in $\G$, $\z^g \leftarrow \text{prox}_{\frac{\lambda\eta_g}{\gamma}\|.\|}[\w_g-\frac{1}{\gamma}\nu^g]$.
   \item Take a gradient ascent step on $\L$ with respect to the $\nu^g$'s: $\nu^g \leftarrow \nu^g + \gamma( \z^g - \w_g )$.
  \item Go back to step~\ref{admm:step1}.
\end{enumerate}
Such a procedure is guaranteed to converge to the desired solution for all
value of $\gamma  > 0$ (however, tuning $\gamma$ can greatly influence the
convergence speed), but solving efficiently step~\ref{admm:step1} can be difficult.
To cope with this issues, several strategies have been proposed in Ref.~\citenum{mairal11}. For simplicity, we do not provide all the details here and refer the reader to Ref.~\citenum{mairal11} for more details.

\section{EXPERIMENTS WITH STRUCTURED DICTIONARIES}\label{sec:exp}
We present here two experiments from Refs.~\citenum{jenatton4}
and~\citenum{mairal11} on learning structured dictionaries, one with a
hierarchical structure, one where the dictionary elements are organized
on a 2D-grid\cite{kavukcuoglu2,hyvarinen2}.
In both experiments, we consider the dictionary learning formulation of Eq.~(\ref{eq:dict}), with a structured sparsity-inducing regularization for the function~$\psi$.
\subsection{Hierarchical Case}
 We extracted patches from the Berkeley segmentation database of natural
images \citep{Martin2001}, which contains a high diversity of scenes.
All the patches are centered (we remove the DC component) and normalized
to have unit $\ell_2$-norm.

\begin{figure}[h]
 \centering
    \includegraphics[width=0.8\linewidth]{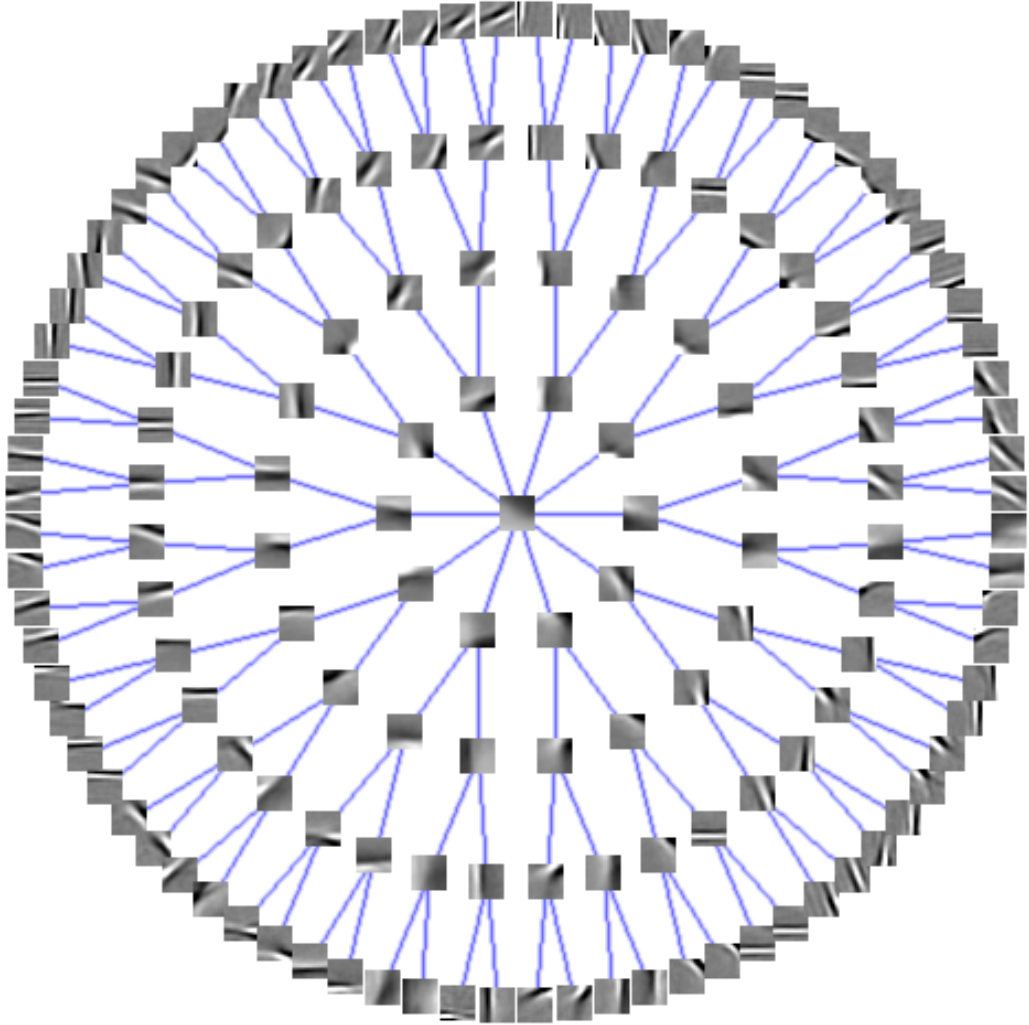}
    \caption{Learned dictionary with a tree structure of depth $5$. The
root of the tree is in the middle of the figure.  The branching factors at depths~$1,2,3,4$ are
respectively $10$, $2$, $2$, $2$. The dictionary is learned on
$50,000$ patches of size $16 \times 16$ pixels. Image taken from Ref.~\citenum{jenatton4}.} \label{fig:tree2}
 \end{figure}

We present visual results on Figures~\ref{fig:tree} and~\ref{fig:tree2},
for different patch sizes and different group structures. For simplicity,
the weights~$\eta^g$ in Eq.~(\ref{eq:def_omega}) are chosen equal to one,
and we choose a penalty~$\psi$ which is a sum of~$\ell_\infty$-norms.  We
solve the sparse decomposition problems~(\ref{eq:lasso}) using the
proximal gradient method of Section~\ref{subsec:optim1}, and use an
alternate minimization scheme to learn the dictionary, as explained in
Section~\ref{subsec:dict}.  The regularization parameter~$\lambda$ is
chosen manually.  Dictionary elements naturally organize in groups of
patches, often with low frequencies near the root of the tree,
and high frequencies near the leaves. We also observe clear correlations
between each parent node and their children in the tree, where children often look like
their parent, but sharper and with small variations.

This is of course a simple visual interpretation, which is
intriguing, but which does not show that such a hierarchical dictionary
can be useful for solving real problems. Some quantitative results can
however be found in Ref.~\citenum{jenatton4}, with an inpainting
experiment of natural image patches.\footnote{In this experiment, the
patches do not overlap. Thus, this experiment does not study the
reconstruction of a full images, where the patches usually
overlap~\cite{elad,mairal8}} The conclusion of this experiment is that to
reconstruct individual patches, hierarchical structures are helpful when
there is a significant amount of noise. 
 \begin{figure}[h]
 \centering
    \includegraphics[width=0.5\linewidth]{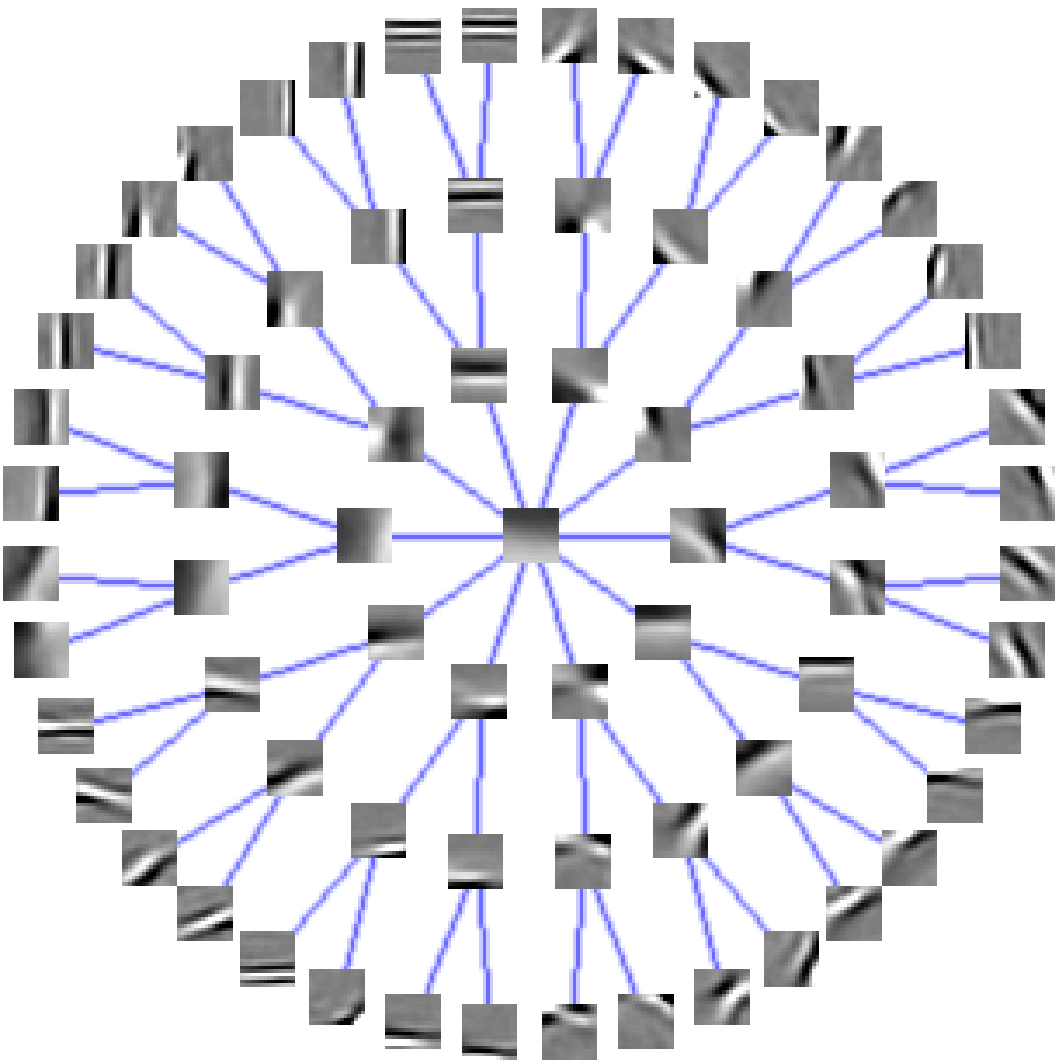}
   \caption{Learned dictionary with a tree structure of depth $4$. The root of the tree is in the middle of the figure.
   The branching factors at depths $1,2,3$ are respectively $10$, $2$, $2$. The dictionary is learned on $50,000$ patches of size $16 \times 16$ pixels. Image taken from Ref.~\citenum{jenatton4}.}
 \end{figure}
\subsection{Topographic Dictionary Learning}
In this experiment, we consider a database of $n=100\,000$ natural image
patches of size $m=12 \times 12$ pixels, for dictionaries of size
$p=400$. 
As done in the context of
independent component analysis (ICA)\cite{hyvarinen2} the dictionary
elements are arranged on a
 two-dimensional grid, and we consider spatial dependencies between them. 
When learned on whitened natural image patches, 
this model called topographic ICA exhibits ``Gabor-like'' functions smoothly organized on the
grid, which the authors call a topographic map.
 As shown in Ref.~\citenum{kavukcuoglu2}, such a result can be
 reproduced with a dictionary learning formulation, using a structured norm for $\psi$.  Following their formulation, we organize the $p$ dictionary elements
 on a $\sqrt{p} \times \sqrt{p}$ grid, and consider $p$
 overlapping groups that are $3 \times 3$ or $4 \times 4$ spatial neighborhoods
 on the grid (to avoid boundary effects, we assume the grid to be cyclic). We
 define $\Omega$ as a sum of $\ell_2$-norms over these groups, since the
 $\ell_\infty$-norm has proven to be less adapted for this task.  Another
 formulation achieving a similar effect was also proposed in Ref.~\citenum{garrigues}
 in the context of sparse coding with a probabilistic model.
 
\begin{figure}
\centering
  \includegraphics[width=0.6\linewidth]{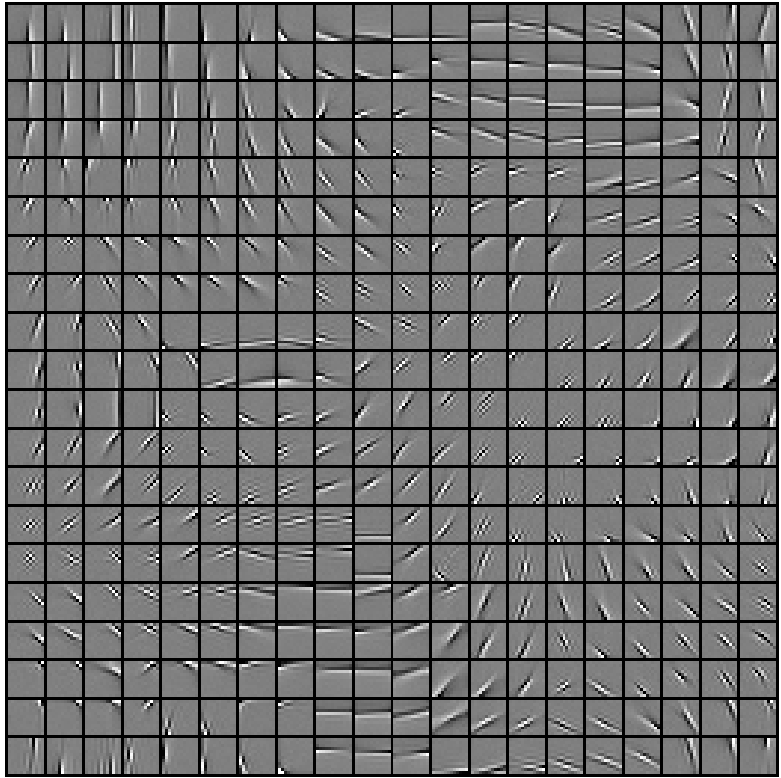} 
\caption{Topographic dictionaries with $400$ elements, learned on a database of $12 \times 12$ whitened natural image patches with $3 \times 3$ cyclic overlapping groups. Image taken from Ref.~\citenum{mairal11}}
\label{fig:topodict}
\end{figure}

\begin{figure}
\centering
  \includegraphics[width=0.6\linewidth]{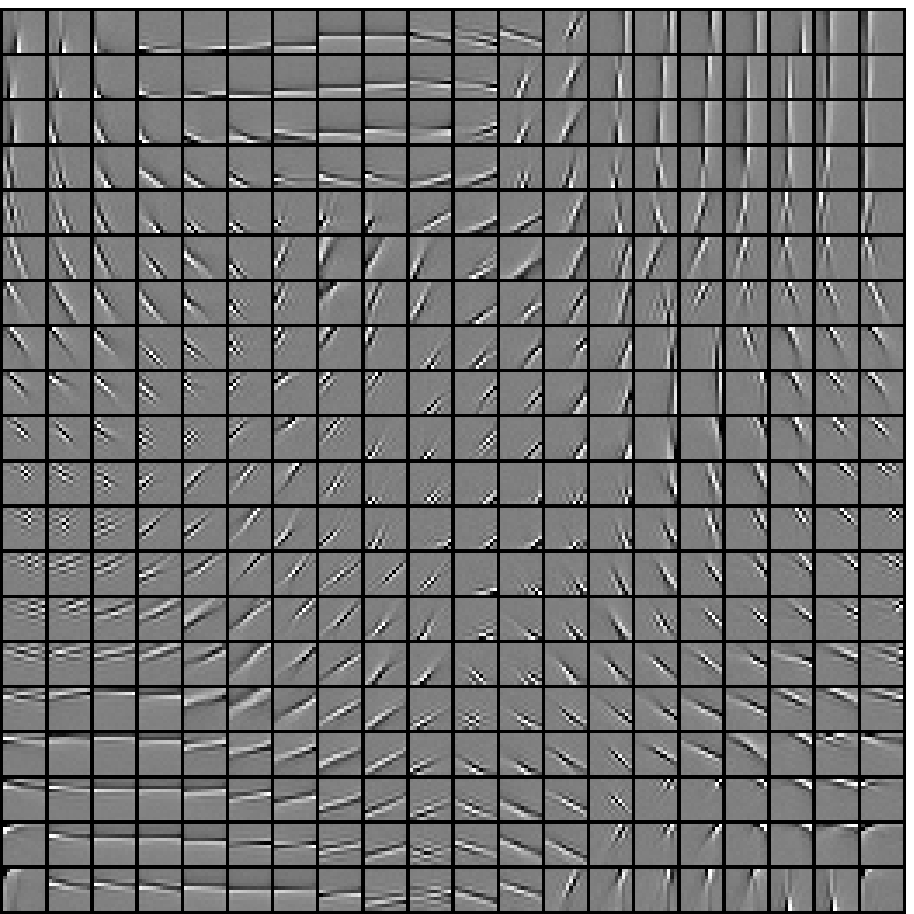} 
\caption{Topographic dictionaries with $400$ elements, learned on a database of $12 \times 12$ whitened natural image patches with $4 \times 4$ cyclic overlapping groups. Image taken from Ref.~\citenum{mairal11}.}
\label{fig:topodict2}
\end{figure}

 As presented in Section~\ref{subsec:dict}, we consider a projected
stochastic gradient descent
 algorithm for learning $\D$---that is, at iteration $t$, we randomly
 draw one
 signal~$\y^t$ from the database $\Y$, compute a sparse code $\w^t$ which
is a solution of Eq.~(\ref{eq:lasso}), and use the update rule
of Eq.~(\ref{eq:sgd}).
In practice, to further improve the
 performance, we use a mini-batch, drawing $500$ signals at each iteration
 instead of one~\cite{mairal7}. This approach mainly differs
from Ref.~\citenum{kavukcuoglu2} in the way the sparse codes $\w^t$ are obtained.
Whereas~Ref.~\citenum{kavukcuoglu2} uses a subgradient descent algorithm to solve them,
 we use the augmenting Lagrangian techniques presented in Section~\ref{subsec:optim2}.
 The natural image patches we use are also preprocessed: They are first centered by
 removing their mean value, called DC component in the image processing literature, and whitened, as 
 often done in the literature~\cite{hyvarinen2,garrigues}.
 The parameter~$\lambda$ is chosen such that in average $\|\y^i-\X\w^i\|_2
 \approx 0.4\|\y^i\|_2$ for every new patch considered by the algorithm, which yields 
 visually interesting dictionaries.
 Examples of obtained results are shown on Figure~\ref{fig:topodict} and~\ref{fig:topodict2}, 
 and exhibit similarities with the maps of topographic ICA\cite{hyvarinen2}.
 
\section{CONCLUSION}
We have presented in this paper different convex penalties inducing both
sparsity and a particular structure in the solution of an inverse
problem. Whereas their most natural application is to model the structure of non-zero patterns of parameter vectors
of a problem, associated for instance to physical constraints in
bio-informatics, neuroscience, they also constitute a natural framework
for learning structured dictionaries. We for instance observe that given 
an arbitrary structure, the dictionary elements can self-organize 
to adapt to the structure. The results obtained 
when applying these methods to natural image patches are intriguing, 
similarly as the ones produced by topographic ICA~\cite{hyvarinen2}.

%%-----------------------------------------------------------
\acknowledgments     %>>>> equivalent to \section*{ACKNOWLEDGMENTS}       
This paper was partially supported by grants from the Agence Nationale de la
Recherche (MGA Project) and from the European Research Council (SIERRA
Project).  In addition, Julien Mairal was supported by the NSF grant
SES-0835531 and NSF award CCF-0939370, and would like to thank Ivan Selesnick
and Onur Guleryuz for inviting him to present this work at the SPIE
conference on Wavelets and Sparsity XIV.

%%-----------------------------------------------------------
%%%%%%%%%%%%%%%%%%%%%%%%%%%%%%%%%%%%%%%%%%%%%%%%%%%%%%%%%%%%%
%%%%%%%%%%%%%%%%%%%%%%%%%%%%%%%%%%%%%%%%%%%%%%%%%%%%%%%%%%%%%
%%%%% References %%%%%

\bibliography{main}   %>>>> bibliography data in report.bib
\bibliographystyle{spiebib}   %>>>> makes bibtex use spiebib.bst

\end{document}